\documentclass[letterpaper]{article} 
\usepackage{aaai2026}  
\usepackage{times}  
\usepackage{helvet}  
\usepackage{courier}  
\usepackage[hyphens]{url}  
\usepackage{graphicx} 
\usepackage{natbib}  
\usepackage{caption} 
\usepackage{algorithm}
\usepackage{algorithmic}
\usepackage{xcolor}  
\usepackage{booktabs}
\usepackage{multirow}
\usepackage{amsmath}
\usepackage{bm}
\usepackage{amsfonts}
\usepackage{amssymb}
\usepackage{newfloat}
\usepackage{listings}
\DeclareCaptionStyle{ruled}{labelfont=normalfont,labelsep=colon,strut=off} 
\floatstyle{ruled}
\newfloat{listing}{tb}{lst}{}
\floatname{listing}{Listing}
\title{One-Shot Refiner: Boosting Feed-forward \\ Novel View Synthesis via One-Step Diffusion}
\author{
    Yitong Dong\textsuperscript{\rm 1,\rm 2}, 
    Qi Zhang\textsuperscript{\rm 2}, 
    Minchao Jiang\textsuperscript{\rm 2,\rm 3},  
    Zhiqiang Wu\textsuperscript{\rm 2,\rm 4},  
    Qingnan Fan\textsuperscript{\rm 2}, \\
    Ying Feng\textsuperscript{\rm 2},  
    Huaqi Zhang\textsuperscript{\rm 2},  
    Hujun Bao\textsuperscript{\rm 1},  
    Guofeng Zhang\textsuperscript{\rm 1}\thanks{Corresponding author.}
}
\affiliations{
    \textsuperscript{\rm 1}Zhejiang University\\
    \textsuperscript{\rm 2}Hangzhou VIVO Information Technology Co., Ltd\\
    \textsuperscript{\rm 3}Xidian University\\
    \textsuperscript{\rm 4}East China Normal University\\
    dongyitong@zju.edu.cn, nwpuqzhang@gmail.com,
    zhangguofeng@zju.edu.cn
}

\usepackage{bibentry}

\begin{document}

\maketitle

\begin{abstract}
We present a novel framework for high-fidelity novel view synthesis (NVS) from sparse images, addressing key limitations in recent feed-forward 3D Gaussian Splatting (3DGS) methods built on Vision Transformer (ViT) backbones. While ViT-based pipelines offer strong geometric priors, they are often constrained by low-resolution inputs due to computational costs. Moreover, existing generative enhancement methods tend to be 3D-agnostic, resulting in inconsistent structures across views, especially in unseen regions.
To overcome these challenges, we design a Dual-Domain Detail Perception Module, which enables handling high-resolution images without being limited by the ViT backbone, and endows Gaussians with additional features to store high-frequency details. We develop a feature-guided diffusion network, which can preserve high-frequency details during the restoration process. We introduce a unified training strategy that enables joint optimization of the ViT-based geometric backbone and the diffusion-based refinement module. Experiments demonstrate that our method can maintain superior generation quality across multiple datasets.
\end{abstract}

\section{Introduction}
Scene understanding and novel view synthesis have seen rapid progress in recent years~\cite{dong2022pedestrian, zhai2025splatloc, zhai_cvpr25_panogs}, largely driven by the transformative impact of differentiable rendering \cite{mildenhall2021nerf} on digital content creation.
Specifically, 3D Gaussian Splatting (3DGS) \cite{3dgs} has achieved an unprecedented breakthrough in this domain, enabling high-fidelity rendering with remarkable performance. However, this approach faces a critical limitation: 3DGS requires extensive, time-consuming optimization for each individual scene, which serves as a significant bottleneck, prohibiting efficient applications and limiting its generalizability.

\begin{figure}[t]
  \centering
   \includegraphics[width=0.47\textwidth]{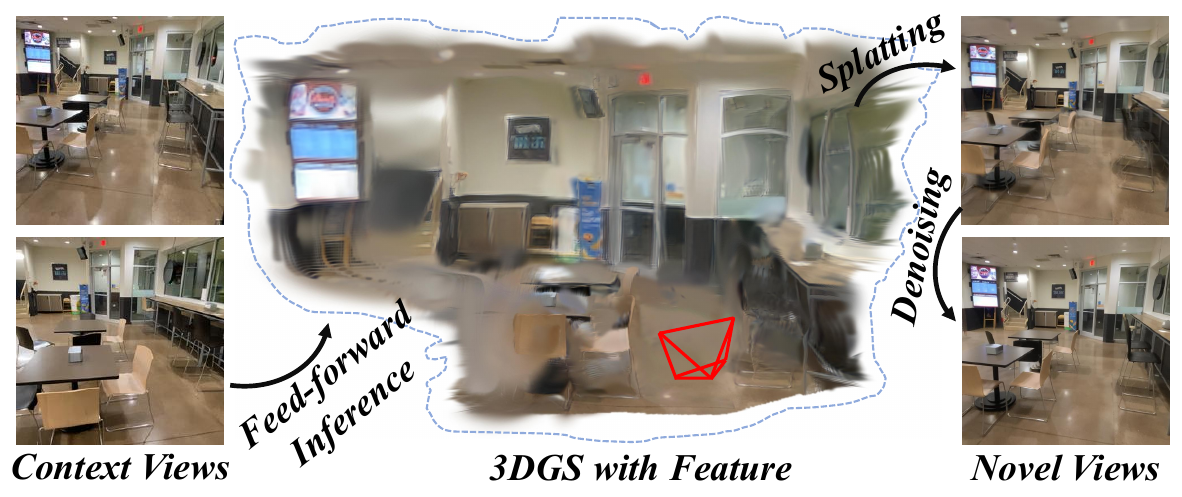}
   \caption{Starting from unposed input images, our method reconstructs 3D Gaussians within a canonical space, and leverages a one-step Stable Diffusion (SD) module to synthesize high-fidelity target views.}
   \label{fig:teaser}
\end{figure}

To overcome this limitation, a new paradigm has emerged built upon 3D foundation models \cite{wang2024dust3r, mast3r, vggt}, which typically leverage a pre-trained Vision Transformer (ViT) backbone. By integrating these powerful 3D geometric priors with 3DGS, such methods \cite{noposplat, jiang2025anysplat} can synthesize novel views from unposed sparse views via rapid single-pass inference. Although this advancement significantly boosts efficiency and generalization, a notable quality gap persists. These feed-forward models often struggle to generate high-fidelity novel views, particularly in rendering high-frequency details. A straightforward strategy might be to feed high-resolution images directly into a larger ViT backbone. However, this approach results in substantial memory overhead and restricts the network’s applicability in real-world scenarios. It is essential to propose an efficient module to enhance the capability of ViT-based backbones to infer high-grained features more effectively.

\begin{figure}[t]
  \centering
   \includegraphics[width=1\linewidth]{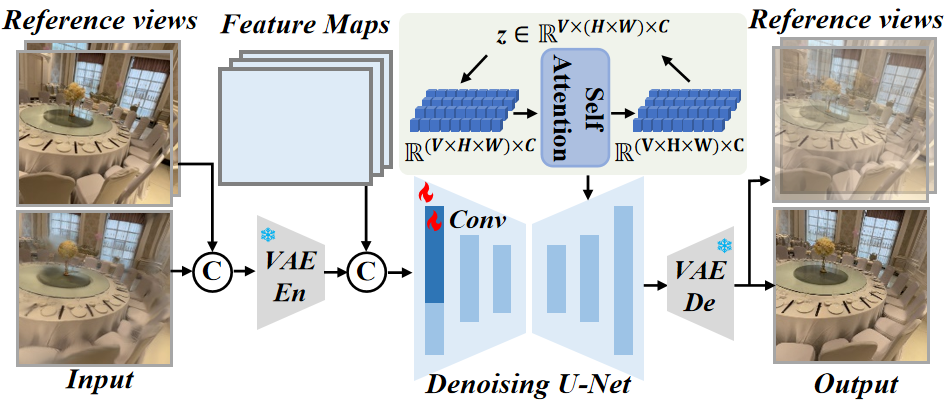}
   \caption{Structure of the diffusion model in NVS.}
   \label{fig:sdmodel}
\end{figure}

Moreover, 2D generative models, such as diffusion models, excel in capturing intricate details and textures, enabling them to produce high-quality 2D images. However, their direct application as a post-processing step for NVS presents a fundamental conflict: these models are 3D-agnostic, resulting in inconsistent structures across views. Compounding this issue is their notoriously slow, iterative sampling process. Our work overcomes these hurdles with a carefully designed Feature-Guided One-Step Diffusion architecture that is both fast and geometrically aware. The core of our innovation, however, lies in how we guide the generation to respect the scene structure. 
As illustrated in Fig.~\ref{fig:sdmodel}, we employ a two-pronged conditioning strategy. A dedicated guidance branch relays explicit geometric priors from the 3D backbone, anchoring the generative process to the scene’s true structure. Simultaneously, the input view serves as a reference condition, guaranteeing that the final output preserves fine-grained information. 

\begin{figure*}[tb]
  \centering
  \includegraphics[width=0.96\textwidth]{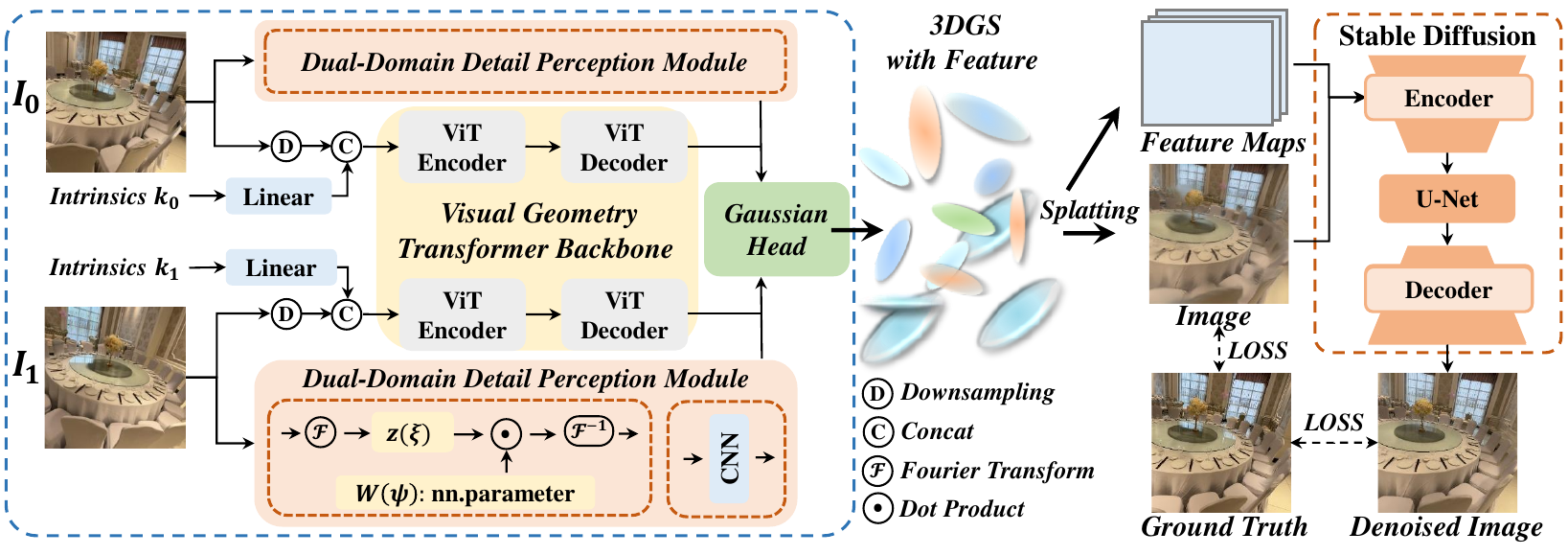}
  \caption{Overview of our pipeline. Starting from a set of unposed images, we first perform spatial downsampling and feed them into a Vision Transformer (ViT)-based backbone for global feature extraction. Simultaneously, we employ a Dual-Domain Detail Perception Module to enhance fine-grained detail perception from both spatial and frequency domains. 
  The fused features are passed into Gaussian Parameter Prediction Heads to directly predict Gaussians with features in a canonical space.
  Finally, a Single-step Denoising module refines these outputs to produce higher-quality novel view synthesis (NVS) results.
  }
  \label{fig:pipeline}
\end{figure*}

Consequently, we achieve an end-to-end framework to enhance the high-quality novel view synthesis from unposed sparse inputs, as shown in Fig. \ref{fig:teaser}.
Specifically, the main contributions of our work are as follows: \begin{enumerate}
\item We design a \textit{Dual-Domain Detail Perception Module} (DD-DPM), 
which enables handling high-resolution images without being limited by the ViT backbone, and endows Gaussians with additional features to store high-frequency details. 
\item We develop a \textit{Feature-Guided One-Step Diffusion} architecture, which can preserve high-frequency details during the restoration process. 
\item We propose an integrated training framework that allows end-to-end training of the ViT reconstruction backbone and the diffusion-based image enhancement module. 
\end{enumerate} 

\section{Related Work}
\noindent\textbf{Radiance fields for novel view synthesis.} 
Neural Radiance Fields (NeRF)~\cite{mildenhall2021nerf} model continuous volumetric density and radiance, enabling high-quality novel view synthesis but relying on positional encoding and importance sampling, which limits real-time performance.
3D Gaussian Splatting (3DGS)~\cite{3dgs} instead represents a scene with millions of anisotropic Gaussians and uses differentiable rasterization to achieve photorealistic rendering at over 30 FPS (1080p). Its efficiency has spurred extensions in rendering~\cite{lu2024scaffold,dong2024global}, surface reconstruction~\cite{huang20242d, chen2024pgsr, yu2024gaussian}, generation~\cite{yi2023gaussiandreamer}, and scene understanding~\cite{qin2024langsplat,jiang2025votesplat}.
Despite these advances, both NeRF and 3DGS rely on dense posed images and require per-scene optimization, limiting their practicality in sparse-view settings.

\noindent\textbf{Generalizable 3D Reconstruction for Sparse View.} 
With the rise of pre-trained foundation models~\cite{wang2024dust3r, mast3r, vggt} that encode strong geometric priors, feed-forward novel view synthesis (NVS) from sparse inputs has received increasing attention.
Unlike other modality-based 3D reconstruction methods~\cite{zhang2024flexcad,zhanggeocad}, these approaches~\cite{zhang2025flare, noposplat, jiang2025anysplat} generally leverage foundation models as backbones to extract geometric cues and improve reconstruction quality.
MuRF~\cite{xu2024murf} aggregates multi-view information through cost-volume construction, while pixelSplat~\cite{charatan2024pixelsplat} exploits epipolar geometry to achieve more accurate depth estimation.

\noindent\textbf{Diffusion priors for novel view synthesis.}
Incorporating diffusion models into reconstruction tasks has been shown to enhance the quality of novel view generation. ReconFusion~\cite{ReconFusion} leverages 2D diffusion priors to recover high-fidelity NeRF from sparsely sampled input views. DiffusionNeRF~\cite{wynn2023diffusionerf} employs a diffusion model to learn gradients of logarithmic RGBD patch priors, which serve as regularized geometry and color for a scene. Nerfbusters~\cite{warburg2023nerfbusters} utilizes diffusion priors to remove artifacts.
Methods such as ReconX~\cite{reconx}, 3DGS-Enhancer~\cite{dgsenhance}, and Difix3D+~\cite{wu2025difix3d+} require per-scene optimization (e.g., Difix3D+ takes 15–30 minutes per scene) and perform poorly under sparse inputs. MVSplat360~\cite{mvsplat360} directly denoises rendered features via video diffusion, but is time-consuming, whereas LatentSplat~\cite{latentsplat} relies on a lightweight VAE-GAN decoder and does not exploit the generative capabilities of a UNet for denoising.
Our goal is to achieve feed-forward novel view synthesis under sparse viewpoints by using one-step Stable Diffusion for refinement.

\section{Method}
Given unposed sparse-view images ${\{I_i\}}^{N}_{i=1} \in \mathbb{R}^{H \times W \times 3}$ and their intrinsics ${\{k_i\}}^{N}_{i=1} \in \mathbb{R}^{3 \times 3}$, 
Our method learns a feed-forward network to generate 3D Gaussians for novel view synthesis (NVS), with an additional refinement module applied to further improve rendering quality.
The scene can be represented by 3D Gaussian Splatting (3DGS):
$g_j:=\{\mathbf{\mu}_j, \;\mathbf{s_j}, \;\mathbf{q_j}, \;o_j, \;\mathbf{c_j}\}$.
Here, $\mathbf{\mu}_j \in \mathbb{R}^3$ and $o_j \in \mathbb{R}$ denote the Gaussian center and opacity, $\mathbf{s}_j \in \mathbb{R}^3$ and $\mathbf{q}_j \in \mathbb{R}^4$ define the 3D covariance, and $\mathbf{c}_j \in [0,1]^3$ represents RGB color via spherical harmonics coefficients. These primitives allow efficient modeling of 3D geometry and appearance for high-quality novel view synthesis.
The overall pipeline of our method is illustrated in Fig.~\ref{fig:pipeline}.
Our network architecture consists of an encoder, a decoder, Gaussian parameter prediction heads, and a final enhancement module.

\subsection{Geometry Transformer Backbone} \label{sec:Geometry Transformer Backbone}
Given unposed images $I_i$, we use a pretrained Vision Transformer (ViT) module~\cite{mast3r} to acquire the geometric information of the scene.
Following the design in~\cite{mast3r}, the geometry transformer module comprises an encoder and a decoder.

\noindent\textbf{Encoder.} 
We initially patchify each RGB image $I_i$ into sequences of image tokens $t^I_i$.
To enhance the network's capability of perceiving geometric information and thereby further optimize reconstruction quality, we inject the intrinsic parameter information of each image into the model.
Specifically, we inject camera intrinsic parameters $[f_x.f_y,c_x,c_y]$ of each image into a linear layer to obtain global feature tokens $t^C_i$, which are then expanded to corresponding image tokens.
Next, the concatenated tokens from each view are individually input to a ViT encoder, with the encoder employing shared weights across all views.

\noindent\textbf{Decoder.} 
The combined tokens from the encoder are then input into the ViT decoder, where cross-view information interaction is achieved through attention modules, resulting in features that contain global geometric information.
This global feature is then used to estimate 3D scene parameters, such as point clouds and Gaussian parameters.

\subsection{Detail-Aware Scene Reconstruction} 
\label{sec:High-Frequency Preserving Scene Reconstruction}
Due to memory constraints, existing feed-forward networks~\cite{charatan2024pixelsplat, noposplat} based on pre-trained transformer foundation models~\cite{wang2024dust3r, mast3r, vggt} commonly restrict input images to low resolutions such as $(256 \times 256)$. 
However, high-resolution images are readily available in real-world scenarios, and directly feeding their downsampled versions into the Geometry Transformer Backbone can lead to the loss of critical structural and appearance information.
Feeding high-resolution images directly into the ViT backbone, or employing additional complex modules to handle them, results in substantial memory overhead. 
This increases the computational burden and restricts the network's applicability in real-world scenarios.
To address this issue, we propose two methods to enable detail-aware reconstruction. 
First, we introduce a lightweight Dual-Domain Detail Perception Module (DD-DPM), which efficiently extracts informative cues from high-resolution inputs with low memory overhead and feeds them into the Gaussian Parameter Prediction Heads.
Second, we extend the feature dimensionality of each Gaussian and implicitly inject information from neighboring pixels in the high-resolution image into the corresponding Gaussian features, thus enhancing the preservation of fine-grained details.

\noindent\textbf{Dual-Domain Detail Perception Module.}
To better capture both local textures and global patterns, we propose a Dual-Domain Detail Perception Module (DD-DPM) that processes image data jointly in the spatial and frequency domains.
Converting images into the frequency domain is a widely adopted strategy, particularly in image super-resolution tasks, as frequency components encode global structures without relying on a strict spatial correspondence.
This property enables more flexible and effective modeling of image details.
Motivated by this, we incorporate a frequency-domain adaptation module to enhance the expressiveness of the extracted features. 
The module begins by applying a 2D Fourier transform to obtain spectral representations of the input images.
\begin{equation}
z(\xi) = \mathcal{F}(I_i),
\end{equation}
where $\mathcal{F}$ denotes the Fourier transforms and $\xi$ is the frequency domain variables.
To focus on the most informative spectral signals, we use an MLP to predict importance scores over normalized frequency coordinates. 
A top-k selection is applied to retain key components, which are then modulated by learned complex weights.
The final output is transformed back through an inverse Fourier transform:
\begin{equation}
F_i' = \mathcal{F}^{-1}(z'(\xi)),
\end{equation}
where $\mathcal{F}^{-1}$ denotes the inverse Fourier transforms.
On the other hand, a lightweight CNN is employed to capture fine-grained texture information from the spatial domain, which is then combined with frequency-domain features and fed into the Gaussian Parameter Prediction Heads.

\noindent\textbf{Gaussian Parameter Prediction Heads.} 
We design heads based on the DPT decoder~\cite{ranftl2021vision} to predict Gaussian parameters. 
Given that the features derived from the ViT encoder and decoder already contain strong geometric priors, we directly utilize these global features through a depth head to obtain depth values, which serve as the center of each Gaussian $\mathbf{\mu}_j$.
This can largely ensure the spatial consistency of Gaussians.

We utilize a second head to predict the remaining Gaussian parameters. To preserve as much valid information contained in high-resolution images as possible, we have designed two strategies:
First, in addition to the global features obtained from the ViT Transformer and the images themselves, we fuse the features derived from the Weighted Fourier Neural Operator, thereby further enhancing the representational capability of Gaussians.
Second, we assign additional features to each Gaussian to contain detailed information, which in turn assists our enhancement module.

\noindent\textbf{Rendering novel view images with feature.}
We design feature-augmented 3D Gaussians, where each Gaussian is equipped with additional features to preserve high-frequency details.
With this design, once the ViT backbone reconstructs the scene and generates the 3D Gaussians, novel-view rendering produces not only RGB color values but also the corresponding high-dimensional Gaussian features. These features can subsequently be used to support 3D-consistent image enhancement in downstream modules.

\subsection{Boosting NVS with Diffusion module} \label{sec:sd}
While existing feed-forward methods benefit from the strong representation capability of ViT models to recover accurate geometric structures, the subsequent novel view rendering process still inevitably encounters 3D artifacts and blurred details.
This problem largely arises from limited supervision provided by the input views, ultimately reducing the quality and perceptual fidelity of the generated novel views.
Therefore, a super-resolution algorithm is needed to improve the fidelity and realism of the rendered images.
To overcome this limitation, we utilize a one-step Stable Diffusion (SD) architecture to enhance the synthesized novel views, yielding higher-quality target images.
Building on this foundation, we further optimize the SD module by incorporating Gaussian features as auxiliary inputs within UNet structure and integrating reference images into UNet’s self-attention module to facilitate inter-image information exchange.

\begin{figure*}[t]
  \centering
  \includegraphics[width=0.98\textwidth]{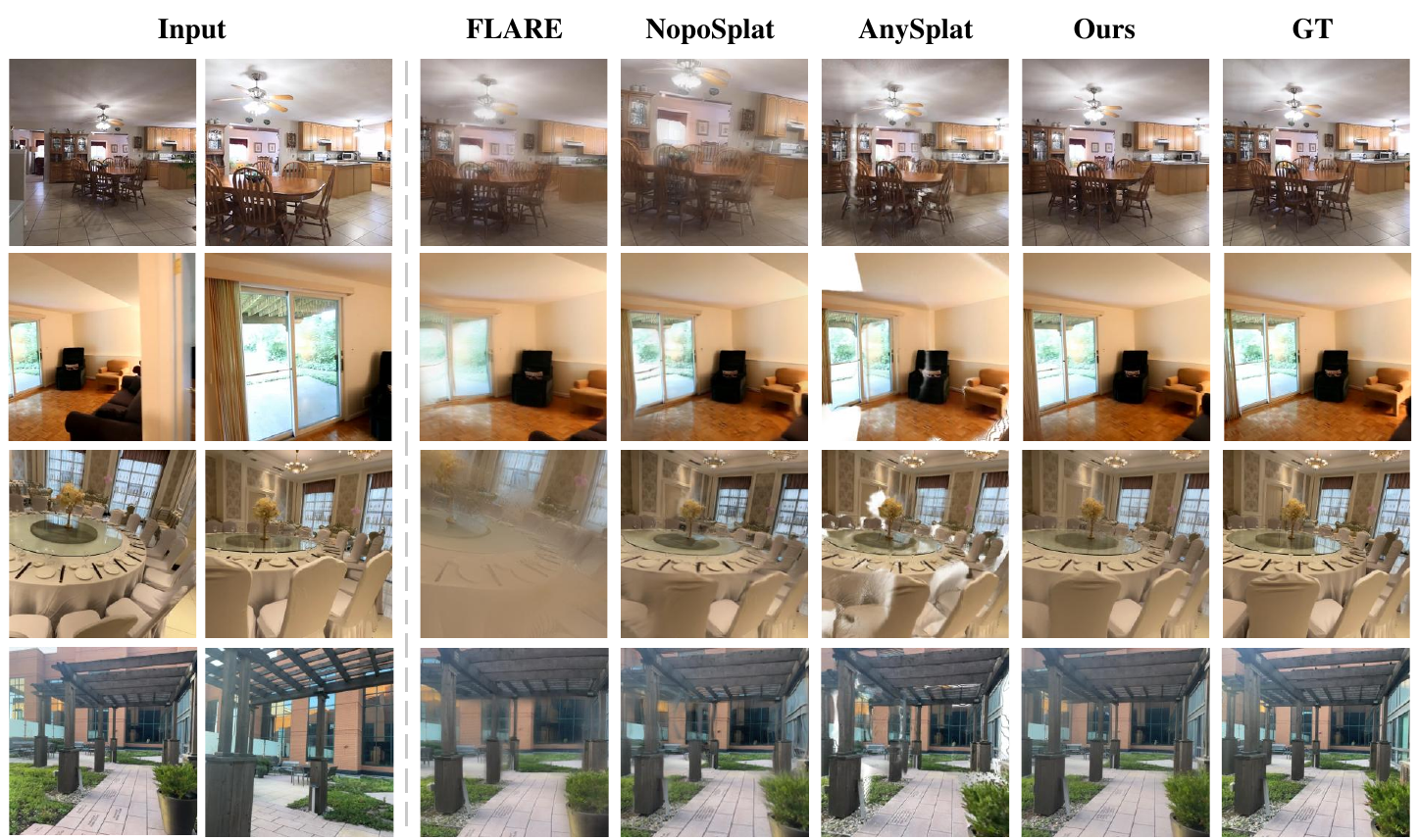}
  \caption{Qualitative comparison on DL3DV and RE10K datasets, all evaluated at a uniform resolution of $\mathbf{512 \times 512}$. Compared with other methods, our approach is capable of recovering finer texture details. 
  }
  \label{fig:dl3dv512}
\end{figure*}

\begin{table*}[t]
  \centering
  \begin{tabular}{l|l|ccc|cccc|c}
  \toprule
  \multirow{2}{*}{Dataset} & \multirow{2}{*}{Methods} & \multicolumn{3}{c|}{Full-Reference} & \multicolumn{4}{c|}{No-Reference} & Perceptual Quality \\
                           &                          & PSNR$\uparrow$ & SSIM$\uparrow$ & LPIPS$\downarrow$ & NIQE$\downarrow$ & MUSIQ$\uparrow$ & M-IQA$\uparrow$ & C-IQA$\uparrow$ & FID$\downarrow$ \\
  \midrule
  \multirow{4}{*}{DL3DV}   & NopoSplat                & 15.53 & 0.47 & 0.57 & 4.94 & 59.77 & 0.51 & 0.29 & 118.42 \\
                           & AnySplat                 & \underline{18.27} &  \underline{0.55} &  \underline{0.27} & \textbf{3.53} &  \underline{64.37} &  \underline{0.68} &  \underline{0.36} &  \underline{64.91} \\
                           & FLARE                    & 15.90 & 0.48 & 0.55 & 4.41 & 59.73 & 0.54 & 0.31 & 122.24 \\
                           & Ours                     &  \textbf{22.67} & \textbf{0.69} & \textbf{0.16} & \underline{3.87} & \textbf{72.86} & \textbf{0.72} & \textbf{0.49} &  \textbf{40.46} \\
  \midrule
  \multirow{4}{*}{RE10K}   & NopoSplat                & 15.81 & 0.60 & 0.54 & 5.85 & 53.97 & 0.49 & \underline{0.31} & 79.81 \\
                           & AnySplat                 & \underline{19.07} & \underline{0.67} & \underline{0.23} & \textbf{4.31} & \underline{60.96} & \underline{0.65} & 0.29 & \underline{42.20} \\
                           & FLARE                    & 16.87 & 0.62 & 0.46 & 5.39 & 52.83 & 0.52 & 0.28 & 67.87 \\
                           & Ours                     & \textbf{20.67} & \textbf{0.70} &  \textbf{0.21}&  \underline{4.71}&  \textbf{69.05}&  \textbf{0.68}&  \textbf{0.39}& \textbf{33.52}  \\
  \bottomrule
  \end{tabular}
  \caption{Novel view synthesis performance on DL3DV and RE10K datasets, all evaluated at a uniform resolution of $\mathbf{512 \times 512}$.}
  \label{tab:dl3dv}
\end{table*}

\begin{table}[t]
\centering
{
\begin{tabular}{l|c|ccc}
    \toprule
    {Methods}
    & LPIPS$\downarrow$ & MS$\uparrow$ & SC$\uparrow$ & BC$\uparrow$ \\
    \midrule 
    MVSplat360   & 0.35 &  \textbf{0.95} & \underline{0.90} &  \underline{0.92} \\
    LatentSplat    & \underline{0.27} &\textbf{0.95} & \underline{0.90}&  0.91 \\
    Ours       &  \textbf{0.19} & \textbf{0.95}  & \textbf{0.92} & \textbf{0.93} \\
  \bottomrule
\end{tabular}
}
\caption{Quantitative evaluation of the 3D consistency performance of generated videos on the DL3DV dataset.}
\label{table:3d}
\end{table}

\begin{table}[t]
\centering
{
\begin{tabular}{l|ccc|c}
    \toprule
    {Methods}
    &PSNR$\uparrow$ & SSIM$\uparrow$ & LPIPS$\downarrow$ & FID$\downarrow$ \\
    \midrule 
    MARINER   & 19.99 & 0.64 & 0.20 & 63.04 \\
    Difix3d+    & \underline{21.67} & \underline{0.67} & \underline{0.18} & \underline{50.30} \\
    Ours       &  \textbf{22.67} & \textbf{0.69}  & \textbf{0.16} & \textbf{40.46} \\
  \bottomrule
\end{tabular}
}
\caption{Quantitative comparison with enhancement methods (MARINER and the SD model used in Difix3D+).}
\label{table:enhance}
\end{table}

\begin{table}[t]
\centering
{
\begin{tabular}{l|ccc|c}
    \toprule
    {Methods}
    & PSNR$\uparrow$ & SSIM$\uparrow$ & LPIPS$\downarrow$ & FID$\downarrow$ \\
    \midrule 
    NopoSplat   & 15.16 & \underline{0.59} & 0.77& 246.80 \\
    AnySplat    & 15.36 & 0.45 & \underline{0.54}& \underline{108.74} \\
    FLARE       & \underline{15.83} & \underline{0.59} & 0.73& 230.59 \\
    Ours        & \textbf{21.64} & \textbf{0.69} & \textbf{0.28}& \textbf{41.74} \\
  \bottomrule
\end{tabular}
}
\caption{NVS performance comparison on the DL3DV dataset at an extended resolution of $\mathbf{1024 \times 1024}$.}
\label{table:1024}
\end{table}

\noindent\textbf{Stable Diffusion.} 
Diffusion Models (DMs) learn data distributions through iterative denoising, and this process becomes significantly more efficient when performed in the latent space using a pre-trained autoencoder~\cite{wu2025omgsr}.
We employ the pre-trained latent diffusion model to mitigate 3D artifacts in novel view synthesis, typically caused by sparse supervision or geometric inconsistencies.
Using the VAE encoder $E_\phi$, latent diffusion network $\epsilon_\phi$, and the VAE decoder $D_\phi$—where $\phi$ denotes the model parameters—the images can be denoised to achieve higher quality.
Building on this, we further incorporate geometric alignment with 3D Gaussian features, enabling the SD module to generate a high-fidelity and 3D-consistent denoised target view $I_d$ via latent-space diffusion decoding.

\noindent\textbf{Gaussian Feature Integration in UNet.} 
After encoding the low-quality rendering image $I_r$ through the VAE encoder $E_\phi$ to obtain the latent feature $F_r \in \mathbb{R}^{C \times \frac{H}{8} \times \frac{W}{8}}$, we render Gaussian feature $F_g \in \mathbb{R}^{C' \times \frac{H}{8} \times \frac{W}{8}}$ along the corresponding dimension $(\frac{H}{8},\frac{W}{8})$ and concatenate it with $F_r$ along the channel dimension.
\begin{equation}
F = \mathcal{CAT}(F_r, F_g),
\end{equation}
The fused features are then fed into the UNet module for cross-modal interaction. Notably, we extend the input channels of the UNet’s initial convolution to accommodate the additional features, initializing the extra dimensions to zero to facilitate stable convergence.

\noindent\textbf{Reference Image Interaction via Self-Attention.} 
We utilize reference images as guidance to assist the diffusion model in generating high-quality images~\cite{wu2025difix3d+, videosd}. 
Specifically, we first concatenate the reference images with the Gaussian-rendered target image and feed the combined result into the VAE encoder to obtain latent features.
We then simultaneously extract Gaussian features under both the reference and target views, and integrate them to produce fused features $z \in \mathbb{R}^{V \times C \times H \times W}$.  
Within the latent diffusion model, we modify the self-attention layers to transform the image interactions within the low-quality image $z \in \mathbb{R}^{V \times C \times (H \times W)}$ into a mixed attention mechanism between the low-quality image and the reference image $z \in \mathbb{R}^{C \times (V \times H \times W)}$.
In this way, we can further capture fine-grained details from reference image.

\noindent\textbf{Two-Stage Pipeline.} 
In the first stage, we employ LoRA to fine-tune the latent diffusion network $\epsilon_\phi$ of the Stable Diffusion (SD) module, encouraging it to restore images degraded by 3D rendering artifacts.
To train the model effectively, we construct a paired dataset using DL3DV, comprising low-quality rendered images and their corresponding high-quality ground-truth (GT) images.
Specifically, following the sampling strategy of~\cite{noposplat}, we encode sparse input views into 3D Gaussian scene representations using a geometry Transformer backbone and scene reconstruction module, and render novel views based on sampled camera extrinsics.  
These rendered images and their GT counterparts form the supervision pairs for training.

In the second stage, directly optimizing rendered images with the Stable Diffusion (SD) architecture can mitigate artifacts to some extent, yet it struggles to preserve fine-grained textures in high-resolution outputs and maintain geometric consistency.
Gaussian features, however, contain both rich textural details and geometric cues.
To address this limitation, we design a pipeline that jointly performs scene reconstruction and integrates a feature-guided SD module to achieve higher-fidelity image synthesis.

\subsection{Training} \label{sec:Training}
The architecture comprises a front-end encoder, a decoder, and a back-end diffusion-based enhancement module. 
The network is fully trainable in an end-to-end manner.
Given the pose, the 3D Gaussian constructed by the front-end is capable of rendering images $I_r$ and their corresponding features. 
These images and features are fed into the enhancement module, which then leverages the priors of diffusion to enhance image quality, resulting in the final images $I_d$.

\noindent\textbf{Training Loss.}
For reconstruction, we use the standard MSE loss together with the perceptual LPIPS loss between the rendered image $I_r$ and the ground-truth image $\hat{I}$:
\begin{equation}
 L_r = \lambda1\cdot MSE(I_r,\hat{I})+\lambda2\cdot LPIPS(I_r,\hat{I}),
  \label{eq:loss_r}
\end{equation}

For the enhancement module, we first directly introduce the MSE loss and LPIPS loss between the denoising image $I_d$ and the target image $\hat{I}$:
\begin{equation}
 L_d = \lambda3\cdot MSE(I_d,\hat{I})+\lambda4\cdot LPIPS(I_d,\hat{I}),
  \label{eq:loss_d}
\end{equation}

To enhance perceptual quality, we incorporate a DINOv2 pre-trained network as a feature extractor to guide the training of the GAN~\cite{ganloss}. This encourages the generator to produce outputs that are not only visually realistic but also semantically consistent with the input. The GAN loss is formulated as:
\begin{equation}
    \begin{split}
        {L}_{g}
        = \lambda_5 \mathbb{E}[\log D(\hat{I})]  
        + \lambda_5 \mathbb{E}[\log (1 - D(G(I)))],
    \end{split}
\end{equation}
where $G$ and $D$ are the generator and discriminator using DINOv2 backbone for training, respectively.
The total loss can be expressed as:
\begin{equation}
 L_{total} = \lambda_r \cdot  L_r + \lambda_d \cdot L_d + \lambda_g \cdot {L}_g,
  \label{eq:loss_t}
\end{equation}

\noindent\textbf{Training pipeline.}
We first train the ViT-based 3D reconstruction module independently to obtain a stable 3D representation. Specifically, images resized to 256×256 are input to the ViT backbone to extract global semantic features. Simultaneously, the Dual-Domain Detail Perception Module processes higher-resolution images at 512×512 to capture fine-grained details. Both feature streams are subsequently fused in the prediction heads to reconstruct the 3D Gaussian representation.
Following the rendering of images and features, 
the SD module is first trained independently.
Subsequently, we perform joint training of the SD module guided by the Gaussian features, as detailed in Sec. \ref{sec:sd}.

\section{Experiments}

\subsection{Training Details} \label{sec:Training Details.}
\noindent\textbf{Dataset.}
We trained our model on the 2K subset of the DL3DV dataset and evaluated it on the benchmark subset~\cite{ling2024dl3dv}.
DL3DV is a large-scale 3D scene dataset widely used for NVS, featuring a variety of reflection, transparency, and lighting conditions.
To further assess generalization, we evaluated our approach on the RealEstate10k (RE10K) dataset~\cite{re10k}, which comprises large-scale indoor real estate videos with multi-view images and accurate camera poses. This setup enables testing the robustness of our method in NVS under complex real-world lighting, textures, and geometric variations.

\noindent\textbf{Evaluation Metrics.}
The quality of novel view synthesis is first measured using three standard metrics: PSNR, SSIM, and LPIPS.
In addition, we have incorporated parameter-free metrics, including DISTS, FID, NIQE, MUSIQ, M-IQA, and C-IQA. 
Detailed analyses are provided in Sec.~\ref{sec:Results.}.

\subsection{Results} \label{sec:Results.}
\noindent\textbf{Novel View Synthesis.}
We selected state-of-the-art (SOTA) feed-forward methods to compare the performance of novel view synthesis (NVS) results with our method on the test set of DL3DV, where the resolution of the final output is 512×512.
As shown in Table.~\ref{tab:dl3dv}, our method attains the superior performance among all compared approaches.

\noindent\textbf{3D Consistent.}
To assess multi-view 3D consistency, consecutive frames are sampled as target views to generate videos. We compare our approach with NVS methods that also perform refinement on rendered images, including MVSplat360 and LatentSplat. 
The Motion Smoothness (MS), Subject Consistency (SC), and Background Consistency (BC) metrics from VBench~\cite{huang2023vbench, zheng2025vbench2, huang2024vbench++} are employed to jointly evaluate 3D consistency. 
As shown in Table~\ref{table:3d}, our method demonstrates substantially improved geometric consistency under sparse viewpoints.

\noindent\textbf{Enhancement Module.}
We compare our feature-guided SD enhancement module with existing enhancement methods by substituting it with MARINER~\cite{mariner} and the SD model from Difix3D+~\cite{wu2025difix3d+}. Table~\ref{table:enhance} shows that our approach surpasses these methods, effectively utilizing 2D reference cues while maintaining 3D consistency via Gaussian features.

\begin{figure*}[!t]
  \centering
  \includegraphics[width=0.96\textwidth]{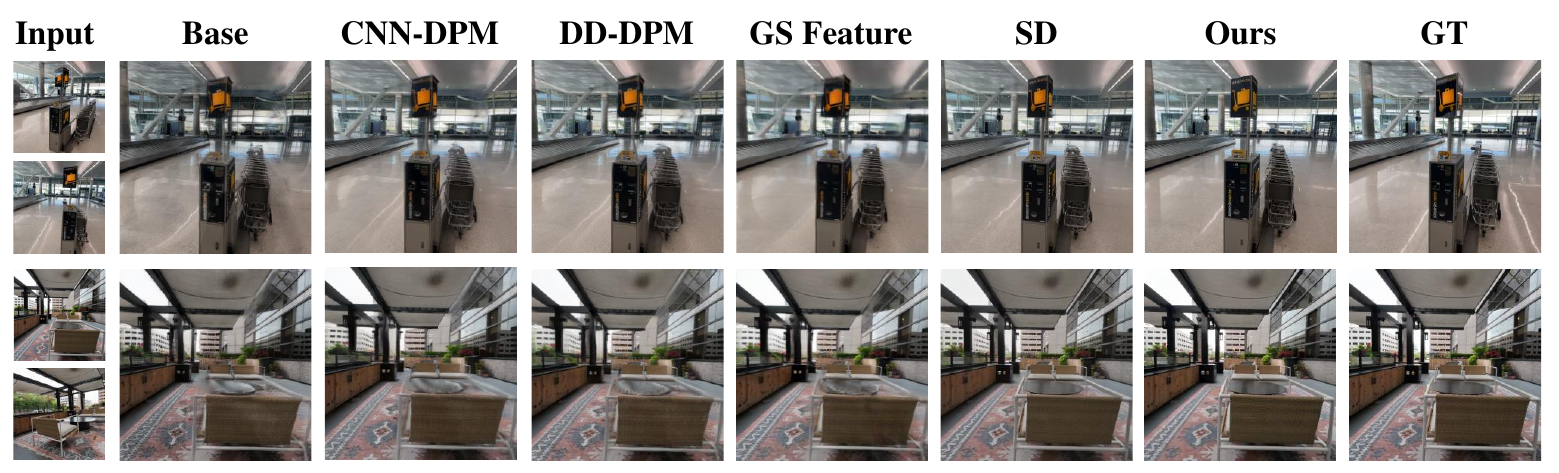}
  \caption{Qualitative ablation results validating the effectiveness of each component. Incorporating Dual-Domain Detail Perception Module (DD-DPM) and the feature-guided SD refine module yields substantially higher-quality novel view synthesis.
  }
  \label{fig:module}
\end{figure*}
\noindent\textbf{Cross-Dataset Generalization.}
We assess the generalization performance of our network on the RE10K dataset without fine-tuning. As shown in Table~\ref{tab:dl3dv}, the results indicate that our method exhibits strong generalization.

\noindent\textbf{Resolution Expansion.}
We expand the output image into a higher-dimensional space without requiring additional training. Specifically, we first render low-resolution images and upsample them to high resolution ($1024\times1024$), followed by generating high-quality outputs using Tile-VAE.

\begin{table*}[t!]
  \centering
  \begin{tabular}{l|ccc|cccc|c}
  \toprule
  \multirow{2}{*}{Methods} & \multicolumn{3}{c|}{Full-Reference} & \multicolumn{4}{c|}{No-Reference} & Perceptual Quality \\
                           & PSNR$\uparrow$ & SSIM$\uparrow$ & LPIPS$\downarrow$ 
                           & NIQE$\downarrow$ & MUSIQ$\uparrow$ & M-IQA$\uparrow$ & C-IQA$\uparrow$ & FID$\downarrow$ \\
  \midrule
   Base &  20.08 & 0.65 & 0.24 &  4.11& 65.98 & 0.55 & 0.29 &  62.19 \\
   +CNN-DPM & 22.10 & 0.69 & 0.23 &4.24 & 64.95 & 0.59 & 0.30 &  62.60\\
    +DD-DPM & 22.14 & 0.69 & 0.22  & 4.15 & 63.47 & 0.60 & 0.30 &  55.32 \\
    +GS Feature(CNN) & \underline{22.22} & \underline{0.69} & 0.27 & 4.93 & 56.15 & 0.55 & 0.30 &  70.51  \\
    +SD & 22.17 & 0.68 & \underline{0.17} & \underline{3.92} & \underline{70.97}  & \underline{0.69} & \underline{0.44} &  \underline{48.58}  \\
    
    +Feature guided SD  & \textbf{22.67} & \textbf{0.69} & \textbf{0.16} & \textbf{3.87} & \textbf{72.86} & \textbf{0.72} & \textbf{0.49} &  \textbf{40.46}  \\
  \bottomrule
  \end{tabular}
  \caption{Ablation study validating the effectiveness of each network component on the DL3DV dataset.}
  \label{tab:ablation2}
\end{table*}

\subsection{Ablation Studies} \label{sec:Ablation Studies.}
We perform ablation studies on a baseline trained with the DL3DV dataset to assess the impact of the proposed module.

\noindent\textbf{Ablation on DD-DPM.}
To evaluate the effectiveness of the Detail-Preserving Module (DPM), we construct two ablation variants:
(1) CNN-DPM, which relies solely on CNN to process high-resolution images and fuses its output with the low-resolution features from the ViT backbone; and
(2) DD-DPM, which additionally introduces a frequency-domain processing branch to enable dual-domain detail perception.
Table~\ref{tab:ablation2} indicates the clear advantage of DD-DPM, emphasizing the importance of frequency-domain features in maintaining fine details.

\noindent\textbf{Ablation on Gaussian Feature.}
To verify the effectiveness of introducing features into Gaussian parameters, we directly connected a simple two-layer CNN structure at the backend, which is used to process the rendered features and images. This lightweight architecture serves as a baseline to assess whether the inclusion of Gaussian feature can bring tangible improvements. By comparing its performance with our full model, we can explicitly quantify the gains from embedding features into Gaussian representations during the rendering and post-processing pipeline.

\noindent\textbf{Ablation on SD Model.}
We constructed a dataset to train the SD network independently and connected it to the backend of the reconstruction module to as an enhancement process for rendered images. This standalone training strategy allows the SD network to specialize in refining details and enhancing visual quality for novel view synthesis, while its integration with the reconstruction module ensures that the enhanced results remain consistent with the geometric structure predicted by the front-end, thereby achieving both accurate novel view synthesis and high-fidelity image quality.

\noindent\textbf{Ablation on Feature-guided SD Refinement Module.}
We introduce a feature-guided SD network and perform joint training of both the ViT-based reconstruction module and the SD network. As illustrated in Table~\ref{tab:ablation2}, incorporating the detailed features rendered by 3D Gaussians into the SD module enables efficient image refinement. This joint optimization strategy ensures that the SD network learns to leverage geometrically consistent detail cues from the rendering process, guiding its refinement efforts toward preserving structural integrity while enhancing visual fidelity, ultimately leading to more coherent and high-quality novel view synthesis compared to standalone enhancement approaches.

\noindent\textbf{Analysis.}
We observe that although incorporating the SD module results in a slight decrease in simple metrics such as PSNR and SSIM, the perceptual quality of the outputs is significantly improved, producing clear and high-quality images.
In contrast, omitting the SD module yields noticeably blurrier results. 
To quantitatively validate this improvement, we evaluate model-based and no-reference metrics including LPIPS, NIQE, MUSIQ, M-IQA, C-IQA, and FID, and find that the model with the SD module consistently outperforms the version without it across all metrics.
Indeed, PSNR can no longer fully capture the true perceptual quality of images. 
LPIPS~\cite{LPIPS} also points out that PSNR and SSIM often do not perfectly align with human perception. Consequently, when differences in simple metrics (e.g., PSNR) are marginal, model-based or no-reference metrics play a more important role in evaluation.

\section{Conclution}
\textbf{Summary.} 
We propose a novel view synthesis method that integrates a feed-forward pipeline with a single-step Stable Diffusion (SD) model. This combination leverages the geometric efficiency of feed-forward methods and the generative strength of SD for detail refinement, producing results with accurate structure and realistic textures.
To enhance consistency between geometry and appearance, we introduce a unified training framework that jointly optimizes geometric representation learning and image generation using features rendered from 3D Gaussians. This tight integration ensures more effective use of geometric features in guiding high-quality image synthesis.
Our framework also paves the way for future work on simplifying parts of the SD architecture, aiming for tighter feature coupling and more efficient view synthesis.
\noindent\textbf{Limitation.} The proposed method does not explicitly model dynamic objects, which limits its applicability in real-world scenarios involving dynamic scenes.

\section*{Acknowledgements}
This work was partially supported by the National Key Research and Development Program of China (No. 2023YFF0905104) and NSF of China (No. 62441222).

\bibliography{aaai2026}

\end{document}